%% file: ms.tex
\documentclass[sigconf,nonacm]{acmart}

\AtBeginDocument{%
  \providecommand\BibTeX{{%
    \normalfont B\kern-0.5em{\scshape i\kern-0.25em b}\kern-0.8em\TeX}}}

\setcopyright{acmcopyright}
\copyrightyear{2022}
\acmYear{2022}
\acmDOI{XXXXXXX.XXXXXXX}

\acmConference[MoDeVVa '22]{19\textsuperscript{th} Workshop on Model Driven Engineering, Verification and Validation}{October 
  2022}{Montreal, Canada}
\acmPrice{15.00}
\acmISBN{978-1-4503-XXXX-X/18/06}

\usepackage[utf8]{inputenc}
\usepackage[T1]{fontenc}
\usepackage{microtype}
\usepackage{amsmath,amsfonts,amsthm}
\usepackage{mathtools}
\usepackage{graphicx}
\usepackage{float}
\usepackage{algorithm}
\usepackage{algpseudocode}
	\algrenewcommand\algorithmicindent{.75em}%
	\algnewcommand{\True}{\textbf{true}}
	\algnewcommand{\False}{\textbf{false}}
	\algnewcommand{\Skip}{\textbf{skip}}
	\algnewcommand{\Break}{\textbf{break}}

\AtEndPreamble{%
\theoremstyle{acmdefinition}

}

\newcommand{\Mult}{\mathit{Mult}}
\newcommand{\TGP}{\mathit{TG}_{\mathrm{P}}}
\newcommand{\TG}{\mathit{TG}}
\newcommand{\PIP}{\mathit{PI}_{\mathrm{P}}}
\newcommand{\PI}{\mathit{PI}}
\newcommand{\CPP}{\mathit{CP}_{\mathrm{P}}}
\newcommand{\CP}{\mathit{CP}}
\newcommand{\GP}{G_{\mathrm{P}}}

\newcommand{\GI}{G_{\mathrm{I}}}
\newcommand{\HI}{H_{\mathrm{I}}}
\newcommand{\msrc}{m_{\mathrm{src}}}
\newcommand{\mtar}{m_{\mathrm{tar}}}
\newcommand{\Gsub}{G_{\mathrm{sub}}}
\newcommand{\Hsub}{H_{\mathrm{sub}}}
\newcommand{\Osub}{O_{\mathrm{sub}}}

\providecommand{\longv}[1]{#1}
\providecommand{\shortv}[1]{}

\begin{document}

\title{A multiplicity-preserving crossover operator on graphs}
\subtitle{Extended version}

\author{Henri Thölke}
\email{Thoelke@students.uni-marburg.de}
\affiliation{%
	\institution{Philipps-Universität Marburg}
	\streetaddress{Hans-Meerwein-Straße 6}
	\city{Marburg}
	\country{Germany}
	\postcode{35043}
}

\author{Jens Kosiol}
\email{kosiolje@mathematik.uni-marburg.de}
\orcid{0000-0003-4733-2777}
\affiliation{%
	\institution{Philipps-Universität Marburg}
	\streetaddress{Hans-Meerwein-Straße 6}
	\city{Marburg}
	\country{Germany}
	\postcode{35043}
}


\begin{abstract}
	Evolutionary algorithms usually explore a search space of solutions by means of \emph{crossover} and \emph{mutation}. 
	While a mutation consists of a small, local modification of a solution, crossover mixes the genetic information of two solutions to compute a new one. 
	For \emph{model-driven optimization} (MDO), where models directly serve as possible solutions (instead of first transforming them into another representation), only recently a generic crossover operator has been developed. 
	Using graphs as a formal foundation for models, we further refine this operator in such a way that additional well-formedness constraints are preserved: 
	We prove that, given two models that satisfy a given set of multiplicity constraints as input, our refined crossover operator computes two new models as output that also satisfy the set of constraints. 
\end{abstract}




\maketitle

\section{Introduction}
\label{sec:introduction}
\input{introduction}

\section{Related work}
\label{sec:related-work}
\input{related-work}

\section{Running example}
\label{sec:running-example}
\input{running-example}

\section{Preliminaries}
\label{sec:preliminaries}
\input{preliminaries}

\section{Introducing secure crossover on graphs}
\label{sec:crossover-operator}
\input{crossover-operator}

\section{Properties of secure crossover}
\label{sec:formal-properties}
\input{formal-properties}

\section{Conclusion}
\label{sec:conclusion}
\input{conclusion}

\begin{acks}
	This work has been partially supported by the \grantsponsor{dfg}{Deutsche Forschungsgemeinschaft (DFG)}{https://www.dfg.de/en/index.jsp}, grant~\grantnum[]{dfg}{TA 294/19-1}.
\end{acks}

\bibliographystyle{ACM-Reference-Format}
\bibliography{literature-short}

\longv{%
	\appendix
	
	\section{Algorithmic details}
	\label{sec:algorithmic-details}
	\input{algorithmic-details}

	\section{Proofs and additional results}
	\label{sec:proofs}
	\input{proofs}
}
\end{document}

%% file: introduction.tex
Model-driven optimization (MDO) performs optimization directly on domain-specific models. 
Early motivation for developing MDO has been that modeling allows for a very expressive representation of optimization problems and their solutions across different domains; at the same time, it allows expressing problems and solutions in domain-specific (modeling) languages, potentially making optimization more accessible for domain experts,~e.g., \cite{BPRKPS12,BP13,HHV15}. 
Furthermore, the explicit representation as a model enables one to leverage domain knowledge to explore the search space more effectively~\cite{HHV15,ZM16}.
While already been used in~\cite{HHV15}, there has been increased awareness recently that MDO makes the search and its employed operators amenable to the application of formal methods: exploiting the formal grounding that graphs, graph transformations, and logics on graphs provide for models, their transformations, and their properties, properties of search operators or the whole search process can be tested or even formally verified~\cite{BZJ21,HSBMZ22,TJK22,JKLT22}. 
 
In practical applications of MDO, often evolutionary algorithms have been used to perform the actual optimization, e.g.,~\cite{FKLW12,AVSNDHH14,FTKWA17,HSBMZ22}. 
In an evolutionary algorithm, normally \emph{crossover} and \emph{mutation} drive the search (see, e.g.,~\cite{ES15}): 
Typically, given an \emph{objective} (or \emph{fitness}) \emph{function} (or a set of these) that formalizes the properties to be optimized, an evolutionary algorithm starts with an \emph{initial population} of randomly generated \emph{solutions}. 
Then, until a \emph{termination condition} is met, a new population is computed from the current one. 
For that, pairs of solutions are randomly selected from the current population, to which crossover is applied with a certain probability; solutions with high fitness have a higher chance of being selected. 
Crossover recombines parts of the two solutions, mixing their genetic information. 
After crossover, the computed \emph{offspring} can additionally be subjected to mutation, where small local changes are performed. 
From the old population and the newly computed solutions the next population is randomly \emph{selected}, again taking the fitness of individual solutions (but possibly also other considerations like maintenance of diversity) into account.
An optimization problem can also be \emph{constrained}, meaning that additional \emph{feasibility constraints} restrict which solutions are considered to be \emph{feasible}. 
In the context of evolutionary algorithms, various approaches have been suggested to deal with feasibility constraints~\cite{Michalewicz95,Coello10}. 
These include immediately discarding infeasible solutions, developing methods to repair them, decrementing their fitness (according to the amount of violation of the feasibility constraints), or designing crossover and mutation operators in such a way that feasibility of solutions is always preserved. 

In MDO, so far, \emph{model transformations} have served as the primary means to perform evolutionary search~\cite{FKLW12,AVSNDHH14,HHV15,FTKWA17,BZJ21,HSBMZ22,JKLT22}. 
But, as we will argue in more detail later (see Sect.~\ref{sec:related-work}), performing crossover directly on models has not been adequately developed yet. 
Recently, a generic crossover operator on graphs, which can serve as a basis for crossover on models, has been introduced~\cite{TJK22}; in the following, we refer to this operator as \emph{generic crossover}. 
When applying generic crossover to two models, it is guaranteed that the computed offspring at least conforms to the structure that is specified by the given meta-model. 
However, meta-models, including those that provide a syntax for the solution of optimization problems, are typically equipped with additional well-formedness (or, in this context, feasibility) constraints. 
The most basic and important of these are \emph{multiplicities}, restricting the allowed incidence relations in instance models. 
The generic crossover operator does not take any feasibility constraints into account, i.e., solutions computed via crossover might violate the given constraints, even if the input solutions satisfied these. 
\emph{In this work, we refine the generic crossover operator from~\cite{TJK22} towards \emph{secure crossover} in such a way that it cannot any longer introduce violations of multiplicity constraints.} 
Practical experience in MDO with evolutionary algorithms that are solely based on mutation as an operator shows that evolutionary search can profit if the used mutation rules cannot introduce new violations of feasibility constraints~\cite{BZJ21,HSBMZ22,JKLT22}. 
Therefore, developing crossover of models in such a way that feasibility is preserved is a promising enterprise. 

The direct contribution of this work is this refinement of an existing crossover operator on graphs in such a way that it preserves the validity of multiplicity constraints. 
By detailedly developing the according algorithms and formally proving their desired properties, we further substantiate the intuition that using models directly as representation of solutions of an optimization problem facilitates the reasoning about search operators and processes. 
It is hardly conceivable to develop a crossover operator that preserves multiplicities when models are encoded, e.g., as bit strings. 
\longv{%
	A few algorithmic details are transferred to Appendix~\ref{sec:algorithmic-details}.
	All proofs (and some additional results) are provided in Appendix~\ref{sec:proofs}.
}
\shortv{In an extended version of this paper~\cite{TK22}, we provide a few additional algorithmic details, all proofs, and some additional results.}

%% file: related-work.tex
In this work, we develop a crossover operator on graphs that preserves multiplicities. 
Generally, this operator refines the one introduced in~\cite{TJK22}; it is also inspired by crossover operators on permutations (see, e.g.,~\cite{Potvin96}) to only compute one offspring. 
In the following, we discuss how crossover has been performed in MDO so far, and we discuss other crossover operators on graphs (or similar structures). 

\paragraph{Crossover in MDO}
In MDO, two encoding approaches have emerged: the \emph{rule-} and the \emph{model-based approach}~\cite{JBBSTZW19}. 
In the rule-based approach (e.g.,~\cite{FKLW12,AVSNDHH14,HHV15,FTKWA17,BFTMW19}), the data structure on which optimization is performed is a sequence of applications of model transformation rules. 
Such sequences also represent models (by applying the sequence to a fixed start model). 
Since the chosen representation is a sequence, classic crossover operators like $k$-point or universal crossover are directly applicable. 
However, crossovers may result in a sequence of transformations that is not applicable to the start model. 
Repair techniques have been suggested to mitigate that problem: inapplicable transformations of a sequence can be discarded~\cite{AVSNDHH14} or replaced by placeholders or random applicable transformations~\cite{BFTMW19}. 
These repairs, however, only serve to obtain a sequence of applicable model transformations. 
Whether or not the model that a sequence of transformations computes satisfies feasibility constraints remains unclear. 

In the model-based approach (e.g.,~\cite{ZM16,BZJ21,HSBMZ22}), optimization is directly performed on models. 
Model transformation rules serve as mutation operators. 
Zschaler and Mandow have suggested that ideas from model differencing and merging could serve as basis for a crossover operator directly applicable to models~\cite{ZM16}. 
However, except for a specific application in which crossover has been based on transformation rules~\cite{HSBMZ22}, evolutionary search on models as search space has been performed via mutation only~\cite{JBBSTZW19,BZJ21,JKLT22}. 
To address this situation, we first developed the mentioned generic crossover operator for graph-like structures~\cite{TJK22}. 
Being generic and based on graphs, it can serve as a formal foundation for crossover operators on models, independently of their semantics or the employed modeling technology. 
However, its application can introduce violations of feasibility constraints. 

In parallel work~\cite{JKT22}, we suggest how to adapt that generic crossover operator to models from the Eclipse Modeling Framework (EMF)~\cite{EMF}. 
We ensure that the results of crossover conform to the general structure of EMF models; however, we do not take additional well-formedness constraints like multiplicities into account. 
In a first small evaluation, evolutionary search can profit from using both crossover and mutation (compared to only using mutation) but only if violations of feasibility constraints that crossover introduced are immediately repaired. 
Crossover hampers search when used without repair. 
These results indicate that directly preserving multiplicities, like we suggest in the current work, is a promising enterprise. 
If possible, directly preserving constraints is a more general approach as ad hoc repair operations might not always be available. 

\paragraph{Crossover on graphs}
There are various approaches to crossover on graphs (or similar structures), more often than not with a specific semantics of the considered graphs in mind. 
Since we address typed graphs and multiplicity constraints, which are both very general concepts, we restrict our comparison to other crossover operators that generally work on graphs and do not depend (too much) on their assumed semantics. 
Crossover operators of which we are aware that can be quite generically applied to graphs are~\cite{GLW00,Niehaus04,MNR10}. 
In~\cite{GLW00}, connectedness properties of a graph are preserved, while in~\cite{Niehaus04} heuristics are used to increase the likelihood of preserving acyclicity or certain reachability properties. 
In~\cite{MNR10}, no preservation of constraints is addressed.
To the best of our knowledge, we are the first to formally ensure the preservation of such complex properties as multiplicities for a crossover operator on graphs.

%% file: running-example.tex
\begin{figure}%
	\centering
	\includegraphics{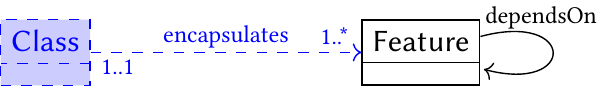}%
	\caption{Metamodel (type graph) for the CRA case}%
	\label{fig:metamodel_CRA}%
\end{figure}

The \emph{class responsibility assignment} (CRA) case is an easy to grasp use case for MDO that is still practically interesting. 
Having been suggested as a contest problem in~\cite{FTW16}, it continues to be used as a problem to illustrate new concepts in MDO with and to benchmark them against~\cite{BZJ21,JKLT22}. 

In the CRA case, a set of \textsf{Features}, representing \textsf{Attributes} and \textsf{Methods}, and their interdependencies are given. 
It is searched for an optimal assignment of the \textsf{Features} to \textsf{Classes}. 
Typically, a solution is considered to be of high quality if its assignment of \textsf{Features} to \textsf{Classes} exhibits low \emph{coupling} between the \textsf{Classes} and high \emph{cohesion} inside of them; for details and a formalization of coupling and cohesion as objective functions we refer to~\cite{FTW16}. 

Fig.~\ref{fig:metamodel_CRA} depicts a metamodel (or type graph) that provides a syntax to represent concrete \emph{problem instances} and \emph{solutions} for the CRA case as models (for brevity, it is considerably simplified compared to the metamodel provided in~\cite{FTW16}). 
The black nodes serve to model problem instances, namely a set of \textsf{Features} and their interdependencies. 
The blue, dashed elements (\textsf{Classes} and their incident edges) serve to extend problem instances to solutions, namely to assign \textsf{Features} to \textsf{Classes}. 
The multiplicities serve to further prescribe which instances of the metamodel are to be considered as \emph{valid} (or \emph{feasible}) solutions. 
In our example they express that every \textsf{Class} should at least \textsf{encapsulate} one \textsf{Feature} ($1..*$) and that every \textsf{Feature} should be encapsulated by exactly one \textsf{Class} ($1..1$).

%% file: preliminaries.tex
In this section we present our formal preliminaries, namely \emph{graphs}, \emph{computation elements with multiplicities}, which serve as a formal basis for models, and the \emph{generic crossover operator} that we refine in this work. 

\begin{definition}[Graph. Graph morphism]
	A \emph{(directed) graph} $G = (V,E,\mathit{src},\allowbreak \mathit{tar})$ consists of finite sets of \emph{nodes} $V$ and \emph{edges} $E$, and \emph{source} and \emph{target functions} $\mathit{src},\mathit{tar}\colon E \to V$. 
	
	Given two graphs $G$ and $H$, a \emph{graph morphism} $f = (f_V,f_E) \colon G \to H$ is a pair of functions $f_V\colon V_G \to V_H$ and $f_E \colon E_G \to E_H$ such that $f_V(\mathit{src}_G(e)) = \mathit{src}_H(f_E(e))$ and $f_V(\mathit{tar}_G(e)) = \mathit{tar}_H(f_E(e))$ for all edges $e \in E_G$.
	A graph morphism is injective/surjective/bijective if both of its components are.
\end{definition}

A typed graph is a graph together with a graph morphism into a fixed type graph. 
Intuitively, the type graph provides the available types for nodes and edges and their allowed incidence relations. 
Using typing to provide graphs with a richer syntax is well-established~\cite{EEPT06}, in particular also when using typed graphs as formal basis for models~\cite{BET12}.
For our application, however, it is advantageous to consider type graphs with more structure: 
Some elements of the type graph serve to model the problem that has to be optimized and others to create a solution. 
Encoding this distinction into the type graph has led to the definition of \emph{computation type graphs} and their \emph{computation graphs}~\cite{TJK22,JKLT22}. 
\emph{Multiplicities} allow expressing certain structural constraints (namely, lower and upper bounds on edge types) that cannot be captured by typing alone; a graph-based formalization is provided in~\cite{TR05}. 
In the following definition of a \emph{computation type graph with multiplicities} and its \emph{computation graphs} we bring together the definition of a type graph with multiplicities with the one of computation (type) graphs. 

\begin{definition}[Computation type graph with multiplicities. Computation graph]
	A \emph{multiplicity} is a pair $[i,j] \in \mathbb{N} \times (\mathbb{N} \cup \{*\})$ such that $i \leq j$ or $j = *$; $\Mult$ denotes the set of all multiplicities. 
	A set $X$ \emph{satisfies} a multiplicity $m = [i,j]$ if $\lvert X \rvert \geq i$ and either $\lvert X \rvert \leq j$ or $j=*$; this satisfaction is denoted as $\lvert X\rvert \in m$. 
	Via $m^{\mathrm{lb}}$ and $m^{\mathrm{ub}}$ we denote the projections to the first or second component. 
	
	A \emph{computation type graph with multiplicities} is a tuple $\overline{\TG} = (\TGP \subseteq \TG,\msrc,\mtar)$, where $\TG$ is a graph, $\TGP$ a designated subgraph, and $\msrc,\mtar\colon E_{\TG} \to \Mult$ are functions, called \emph{edge multiplicity functions}. 
	The graph $\TG$ is called the \emph{type graph}, and $\TGP$ the \emph{problem type graph}. 
	
	A \emph{computation graph} $\overline{G} = (G,t_G)$ over $\overline{\TG}$ is a graph $G$ together with a graph morphism $t_G\colon G \to \TG$. 
	The intersection of $G$ with $\TGP$ (in $\TG$) is called the \emph{problem graph} of $\overline{G}$ and is denoted as $\GP$. 
	A \emph{morphism} $f\colon \overline{G} \to \overline{H}$ between computation graphs $\overline{G} = (G,t_G)$ and $\overline{H} = (H,t_H)$ is a graph morphism $f\colon G \to H$ such that $t_H \circ f = t_G$.
	
	A \emph{computation graph} $\overline{G} = (G,t_G)$ typed over $\TG$ is said to \emph{satisfy the multiplicities} of a computation type graph with multiplicities $(\TGP \subseteq \TG,\msrc,\mtar)$ if 
	\begin{itemize}
		\item for all edges $e \in E_{\TG}$ and all nodes $p \in \{n \in V_G \mid t_G(n) = \mathit{src}_{\TG}(e)\}$ it holds that 
		\begin{equation*}
			\lvert\{e^{\prime} \in E_G \mid t_G(e^{\prime}) = e \text{ and } \mathit{src}_G(e^{\prime}) = p\}\rvert \in \mtar(e)
		\end{equation*}
		and
		\item for all edges $e \in E_{\TG}$ and all nodes $p \in \{n \in V_G \mid t_G(n) = \mathit{tar}_{\TG}(e)\}$ it holds that 
		\begin{equation*}
			\lvert\{e^{\prime} \in E_G \mid t_G(e^{\prime}) = e \text{ and } \mathit{tar}_G(e^{\prime}) = p\}\rvert \in \msrc(e) .
		\end{equation*}
	\end{itemize}
\end{definition}

A computation graph defines a problem instance via its problem graph; (candidate) solutions are all computation graphs with coinciding problem graphs.
\begin{definition}[Problem instance. Search space. Solution]
	Given a computation type graph with multiplicities $\overline{TG} = (\TGP \subseteq \TG,\msrc,\allowbreak \mtar)$, a \emph{problem instance} for $\overline{\TG}$ is a computation graph $\overline{\PI}$ over $\overline{\TG}$. 
	The \emph{search space} $\mathcal{S}(\overline{\PI})$ defined by $\overline{\PI}$ consists of all computation graphs $\overline{G}$ over $\overline{\TG}$ for which there exists (a typing-compatible) isomorphism between the problem graphs $\GP$ and $\PIP$; elements of $\mathcal{S}(\overline{\PI})$ are called \emph{candidate solutions}, or \emph{solutions} for short, for the problem instance $\overline{\PI}$. 
	A solution is \emph{feasible} if it satisfies the multiplicity constraints of $\overline{\TG}$ and \emph{infeasible} otherwise.
\end{definition}

\begin{example}
	Formally, the metamodel for the CRA case depicted in Fig.~\ref{fig:metamodel_CRA} can considered to be a computation type graph with multiplicities; the black elements constitute the problem type graph; the multiplicity functions are defined by the annotations of the edges. 
	Fig.~\ref{fig:example-solutions} presents two feasible computation graphs $\overline{G}$ and $\overline{H}$. 
	The typing is indicated by denoting the nodes with their types; edge types are omitted for brevity (they can unambiguously be inferred). 
	The identifiers \textsf{f1} etc. just serve to be able to speak about the individual elements. 
	Because the problem graphs of $\overline{G}$ and $\overline{H}$ (the black \textsf{Features} plus edge) coincide, they constitute (feasible) solutions for the same problem instance.
	
	\begin{figure}%
		\centering
		\includegraphics{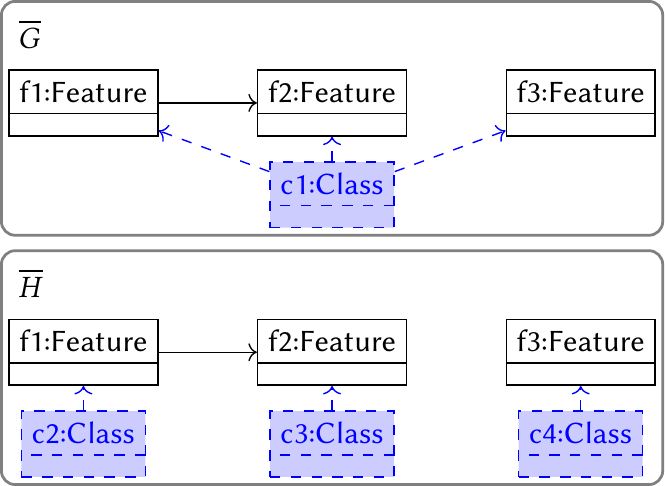}%
		\caption{Two computation graphs for the CRA case that constitute solutions for the same problem instance}%
		\label{fig:example-solutions}%
	\end{figure}
\end{example}

Summarizing, our setting is that a fixed meta-model with multiplicities provides the syntax to model different instances of an optimization problem and their solutions. 
In our example, the general optimization problem is the CRA case and concrete problem instances are different configurations of \textsf{Features} and their interdependencies for which an optimal assignment to \textsf{Classes} has to be found. 
Generally, the problem model of an instance model determines for which problem instance this model is a solution, i.e., to which search space it belongs. 
Evolutionary search, then, is performed for a concrete problem instance, i.e., inside of a fixed search space, which is determined by the isomorphic problem models of its members.
As mentioned in Sects.~\ref{sec:introduction} and \ref{sec:related-work}, it is established to use model transformations (which do not change the problem part) as mutation operators during evolutionary search. 
This paper is concerned with developing a crossover operator that takes two solutions from the same search space as input and computes offspring solutions that belong to the same search space. 
At that the crossover operator shall produce feasible offsprings from feasible input. 
 
As foundation for that, we recall the \emph{generic crossover operator} that has been introduced in~\cite{TJK22}. 
There, crossover is declaratively defined and no constructive algorithm is provided; furthermore, it is defined in an abstract category-theoretic setting. 
Intuitively, that crossover operator prescribes that two input solutions are to be \emph{split} into two parts each and these parts are to be \emph{recombined} crosswise by computing a union.
This results in two new solutions, also called \emph{offspring}. 
A \emph{crossover point} serves to identify elements from the two input solutions, i.e., it designates elements that are only to appear once in the union. 
The next definition specializes generic crossover to computation graphs.

\begin{definition}[Generic crossover]
	Given a computation type graph with multiplicities $\overline{\TG}$, a problem instance $\overline{\PI}$ for it, and two solutions $\overline{G}$ and $\overline{H}$ from $\mathcal{S}(\overline{\PI})$, applying \emph{generic crossover} amounts to the following:
	\begin{enumerate}
		\item \emph{Splitting}: 
		The underlying graph $G$ of $\overline{G}$ is split into two subgraphs $G_1,G_2 \subseteq G$ such that
		\begin{enumerate}
			\item both $G_1$ and $G_2$ contain $\GP$ and
			\item every element of $G$ belongs to either $G_1$ or $G_2$ (or both).
		\end{enumerate}
		$G_1$ and $G_2$ become computation graphs by defining their typing morphisms as the respective restrictions of $t_G$. 
		The intersection $\GI$ of $G_1$ and $G_2$ (in $G$) is called \emph{split point}. 
		The graph $H$ is split in the same way. 
		
		\item \emph{Relating $G$ and $H$ (crossover point)}: 
		A further solution for $\PI$, i.e., a computation graph $\overline{\CP} = (\CP,t_{\CP})$ with problem graph $\CPP$ being isomorphic to $\PIP$ is determined such that $\CP$ can be considered to be a subgraph of both split points $\GI$ and $\HI$. 
		Formally, this means that there are injective morphisms from $\overline{\CP}$ to $\overline{\GI}$ and $\overline{\HI}$. 
		The solution $\overline{\CP}$ (together with the two injective morphisms) is called a \emph{crossover point}.
		
		\item \emph{Recombining}: The underlying graph $G_1H_2$ of the first \emph{offspring solution} $\overline{G_1H_2}$ is computed as the union of $G_1$ and $H_2$ over $\CP$. 
		That is, elements from $G_1$ and $H_2$ that share a preimage in $\CP$ only appear once in the result. 
		A typing morphism $t_{G_1H_2}\colon G_1H_2 \to \TG$ is obtained by combining the ones of $G_1$ and $H_2$; hence, a computation graph $\overline{G_1H_2}$ is computed. 
		A second offspring solution $\overline{G_2H_1}$ is computed analogously by unifying $G_2$ and $H_1$ over $\CP$.
	\end{enumerate}
\end{definition}

The results obtained in~\cite{TJK22} ensure that the two offspring solutions computed in that way are indeed solutions for the given problem model $\overline{\PI}$: their typing morphisms are well-defined and their problem graphs are isomorphic to $\PIP$~\cite[Prop.~2]{TJK22}. 
Moreover, no matter how the splits of the input graphs $G$ and $H$ are chosen, a crossover point can be found; thus, crossover is always applicable~\cite[Lem.~1]{TJK22}. 
Albeit, the computed offspring does not need to satisfy the given multiplicity constraints, even if the inputs $G$ and $H$ and/or the split parts $G_1,G_2,H_1,H_2$ do.  

\begin{example}\label{ex:example-generic-crossover}
	Figure~\ref{fig:example-offspring} depicts two solutions that can result from applying generic crossover to the computation graphs $\overline{G}$ and $\overline{H}$ from Fig.~\ref{fig:example-solutions}. 
	In the corresponding computations, $G$ is split such that $G_1$ does not contain the reference from \textsf{Class c1} to \textsf{Feature f3} and $G_2$ is the whole of $G$.  
	Consequently, $\GI$ coincides with $G_1$. 
	Similarly, $H$ is split by omitting the reference from \textsf{Class c4} to \textsf{Feature f3} in $H_1$ and omitting \textsf{Class c3} together with its reference from $H_2$. 
	Consequently, $\HI$ contains the three \textsf{Classes c2, c3, c4} but only the reference from \textsf{c2} to \textsf{f1}. 
	As crossover point, we choose a graph that, beyond the given problem graph, contains a common preimage for \textsf{Classes c1} and \textsf{c2} and their respective reference to \textsf{Feature f1}, thus identifying them. 
	As can be seen, the resulting offspring $\overline{G_1H_2}$ is desirable as it assigns the independent \textsf{Feature f3} to its own \textsf{Class}. 
	In contrast, offspring $\overline{G_2H_1}$ is infeasible, containing an empty \textsf{Class} (i.e., a \textsf{Class} with no \textsf{Feature}) and a multiply assigned \textsf{Feature}.
	
	\begin{figure}%
		\centering
		\includegraphics{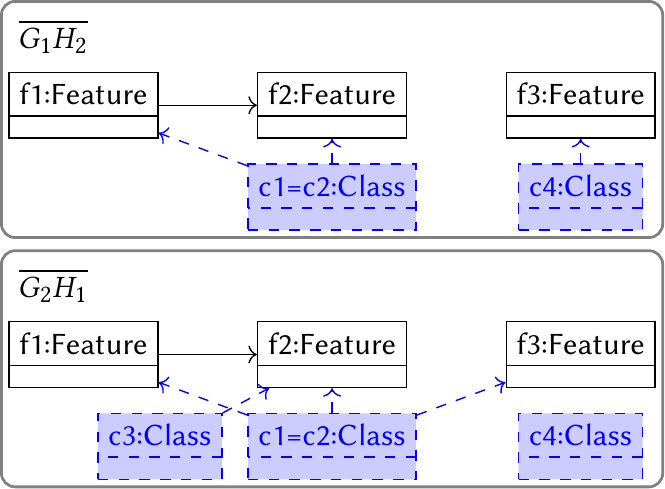}%
		\caption{Two computation graphs resulting as offspring from generic crossover}%
		\label{fig:example-offspring}%
	\end{figure}
\end{example}

We close this section with introducing some abbreviating notations and simplifying assumptions. 
First, we assume the problem graphs of the occurring computation graphs to be identical (and not merely isomorphic). 
Furthermore, from a formal point of view, morphisms (and inclusions) relate the different occurring graphs like split parts, split points, crossover points etc. 
Considering these explicitly leads to quite some notational overhead. 
To avoid that, we use common names of variables and indices to convey the relation between elements. 
For example, we write $x_G$, $x_{G_1}$, $x_{\PIP}$, $x_{\CP}$, etc. to indicate that the node $x_{\CP}$ from crossover point $\CP$ is mapped to the node $x_{\PIP}$ in the problem part, the node $x_{G_1}$ in split part $G_1$ of a computation graph $G$, etc.

%% file: crossover-operator.tex
\input{intro-crossover}

\input{algorithm-crossover-extended}

%% file: intro-crossover.tex
In this section, we introduce our crossover operator, which we call \emph{secure crossover}, and prove its central properties, namely being multiplicity-preserving (\emph{feasibility-preserving}) and still being able to produce any valid graph as offspring (\emph{coverage of search space}). 
As mentioned, we develop our crossover operator as a refinement of the one introduced in~\cite{TJK22} and recalled in Sect.~\ref{sec:preliminaries}.  
Our construction is guided by the following objectives.
\begin{enumerate}
	\item Applied to feasible input graphs, secure crossover shall compute feasible offsprings. 
	
	\item The search space should not get unnecessarily restricted by our operator. 
 
	At least, it should be possible to obtain every feasible graph as a result from crossover. 
	That is, our operator should not cut off feasible regions from the search space. 
	
	\item It should be possible to perform crossover efficiently. 
\end{enumerate}

Finding splits of two graphs, however, and a crossover point such that 
both computed offspring solutions satisfy all multiplicities seems to 
require intensive analysis and becomes very inefficient. Therefore, 
we take inspiration from crossover operators that have been developed 
for the case where permutations (of numbers) are chosen as representation. 
There, to ensure that crossover of two permutations results in a 
permutation again, several crossover operators have been developed 
that only compute one offspring solution (see, e.g.,~\cite{Potvin96} 
for an overview). Analogously, in this work the basic idea is to 
focus entirely on one offspring solution. Computing a single feasible 
offspring requires significantly less computational effort than ensuring 
both offsprings to be feasible. We suspect that it could be cheaper to 
apply the operator we propose in this paper twice, instead of trying 
to compute an application of a crossover operator with two feasible 
offsprings.

Since, for formal reasons, we still want our suggested crossover operator 
to adhere to the definition of generic crossover, we suggest a trivial 
way to compute a second solution (namely, to just unify the whole given 
solution graphs $G$ and $H$ over the crossover point selected to compute 
the first offspring). However, in an implementation one would probably 
just omit this step as the likelihood of this second offspring being 
valuable is very low. Accordingly, in the following we entirely focus on 
the presentation of the computation of the first offspring.

%% file: algorithm-crossover-extended.tex
In all of the following, we assume to be given two computation graphs 
$\overline{G}$ and $\overline{H}$ that are solutions for the same 
fixed problem instance $\overline{\PI}$ (all typed over the same computation 
type graph with multiplicities $\overline{TG}$). We will mainly 
speak about their underlying graphs $G$, $H$, $\PI$, etc. and treat 
their typing implicitly. Recall that $\PIP$ refers to the problem 
graph of $\PI$ that captures the concrete problem instance that is to 
be optimized. 

\subsection{General procedure}

Our general procedure is presented in Algorithm~\ref{alg:secure-crossover} 
and works as follows: We construct a subgraph $\Gsub$ of $G$ as a first 
split part in such a way that $\Gsub$ is always feasible when $G$ is 
(line~1). Knowing $\Gsub$, we construct $\Hsub$ (as a second 
split part of $H$) and a crossover point $\CP$ in parallel, monitoring that applying 
crossover with that data is not going to introduce new violations of 
the multiplicity constraints (lines 2--4). 
Computing the crossover point $\CP$ and $\Hsub$ simultaneously allows for 
a much easier handling of the upper bound of the multiplicities: 
Whenever a node is included into $\CP$, all edges that attach to that
node\rq{}s counterparts in $\Gsub$ and $\Hsub$ will be present in the 
offspring, attached to that node from $\CP$. This can lead to the 
resulting amount of edges being higher than the multiplicity permits.
If we were to compute $\Hsub$ first, we would therefore not be able to 
ensure that a later choice of $\CP$ could not lead to any breach of 
a maximum. These breaches would either need  to be avoided when 
computing $\CP$, which might not always be possible, or we would be 
forced to alter $\Gsub$ and $\Hsub$, creating a lot of additional 
computational effort. This can be avoided by selecting $\CP$ alongside 
the subgraph $\Hsub$. 
Finally, the offspring solution $O_1$ is computed as 
the union of $\Gsub$ and $\Hsub$ over $\CP$ (line~5). As already 
mentioned, to still comply with generic crossover, we formally also 
compute a second offspring $O_2$ as the union of $G$ and $H$ over $\CP$; 
however, we do not expect to perform this computation in practice.

\begin{algorithm}
\caption{\label{alg:secure-crossover}secureCrossover($G,H,\PIP$)}

\begin{algorithmic}[1]
\State{$\Gsub \gets$ createGsub($G,\PIP$)}
\State{global $\CP,\Hsub \gets \PIP$} 
\State{constructCP($\Gsub,H$)}
\State{processFreeNodes($\Gsub,H,\Hsub)$}
\State{$O_1 \gets \Gsub \cup_{\CP} \Hsub$}
\State{$O_2 \gets G \cup_{\CP} H$}
\end{algorithmic}
\end{algorithm}

\subsection{Creating \texorpdfstring{$\Gsub$}{Gsub}}
\label{sec:creation-Gsub}
The creation of $\Gsub$ is detailed in Algorithm~\ref{alg:sichererSplit-1}. 
We initialize $\Gsub$ with $\PIP$ as the problem graph is to be included anyhow. 
Then, for all remaining nodes of $G$, we check whether they are already included in $\Gsub$ (e.g., because they serve as target node for an included egde) and, if not, we randomly decide whether or not the node should be included (line~5). 
For included nodes, we also include adjacent edges (line~8). 
Both nodes and edges are understood to be included into $\Gsub$ with they same type they have in $\overline{G}$. 
Formally, $\Gsub$ becomes a computation graph over $\overline{TG}$ by defining its type function $t_{\Gsub}$ as restriction of $t_G$ to $\Gsub$. 
For simplicity, we leave this implicit in the pseudocode and in all of the following. 

\begin{algorithm}
\caption{\label{alg:sichererSplit-1}createGsub($G,\PIP$)}

\begin{algorithmic}[1]
\State{$\Gsub \gets \PIP$}
\ForAll{$x_G \in V_G$}
\State{bool $\mathit{included} \gets x_{\Gsub} \in V_{\Gsub}$}
\If{$!\mathit{included}$}
\State{$\mathit{included} \gets$ decideInclusion($x_G$)}
\EndIf{}
\If{$\mathit{included}$}
\State{includeAdjacentEdges($x_G,G$)}
\EndIf{}
\EndFor{}
\end{algorithmic}
\end{algorithm}

The inclusion of edges is done by including a random set of the edges of $G$ in $\Gsub$ whose size is between the required lower and upper bound. 
In case it should not be possible to reach the lower bound (because this is violated in $G$) at least all the available edges are included in $\Gsub$.
\longv{We provide an algorithm for this and an extended explanation in Appendix~\ref{sec:algorithmic-details}.} 
\shortv{We provide an algorithm for this and an extended explanation in the extended version of this paper~\cite{TK22}.} 

\begin{example}\label{ex:createGsub}
	In the CRA case, called for a feasible solution $G$, \emph{createGsub} will always return the whole of $G$ as $\Gsub$. 
	All \textsf{Features} are directly included into $\Gsub$ because they are part of the problem graph. 
	Since the incoming \textsf{encapsulates}-edge has a source multiplicity of $[1,1]$, for every \textsf{Feature}-node the call of \emph{includeAdjacentEdges} includes the incoming \textsf{encapsulates}-edge (that exists because of feasibility) and the \textsf{Class} from which it starts. 
	Since, again by feasibility, empty \textsf{Classes} cannot occur, that process results in $G$. 
\end{example}

\subsection{Creating \texorpdfstring{$\Hsub$}{Hsub} and \texorpdfstring{$\CP$}{CP}}
 
The main difficulty in constructing $\Hsub$ and $\CP$ is that, as soon as a node $x_{\CP}$ in the crossover point prescribes that nodes $x_{\Gsub}$ from $\Gsub$ and $x_{\Hsub}$ from $\Hsub$ are to be identified in the offspring, we need to ensure that no upper bound violations can be introduced by that identification: 
Even if in both $\Gsub$ and $\Hsub$ the upper bounds are satisfied, the sum of adjacent edges of a certain type of edges to $x_{\Gsub}$ in $\Gsub$ and $x_{\Hsub}$ in $\Hsub$ might exceed the upper bound of that type of edge. 
Constructing $\Hsub$ and $\CP$ in parallel, we can react to that by either (i) not including too many edges in $\Hsub$, (ii) also including the edges into $\CP$ (which reduces their number in the offspring), or (iii) dropping edges from $\Gsub$. 
As nodes from $\CP$ are the sensitive issue, we start constructing $\Hsub$ and $\CP$ along the nodes we know to appear in them -- the nodes from the problem graph $\PIP$; it actually suffices to start form the \emph{border} of $\PIP$ in $H$, namely the nodes of $\PIP$ which have an adjacent edge (in $H$) to a node that does not belong to $\PIP$. 

\begin{algorithm}
\caption{\label{alg:constructCP}constructCP($\Gsub,H$)}

\begin{algorithmic}[1]
\State{queue $Q \gets$ border($\PIP,H$)}
\While{$Q$.notEmpty()}
\State{$x_H \gets Q$.pop()}
\ForAll{$\mathit{dir} \in \{\mathit{src},\mathit{tar}\}$} 
\ForAll{$e \in E_{\TG}$ with $\mathit{dir}_{\TG}(e)=t_H(x_H)$} 
\ForAll{$e^{\prime}_H \in \mathit{edges}(H,e,\mathit{dir},x_H)$}
\State{$y_H \gets \overline{\mathit{dir}}(e_H)$}
\State{bool $\mathit{included} \gets y_{H} \in V_{\Hsub}$}
\If{$\mathit{included}$}
\State{bool $\mathit{edgeIncluded} \gets e^{\prime}_H \in E_{\Hsub}$}
\If{$\mathit{edgeIncluded}$}
\State{\Skip}
\EndIf{}
\State{bool $\mathit{includedCP} \gets y_{H} \in V_{\CP}$}
\EndIf{}
\If{$!\mathit{included}$}
\State{$\mathit{included} \gets$ decideInclusion($y_H$)}
\EndIf{}
\If{$\mathit{included}$}
\State{$\mathit{edgeIncluded} \gets$ decideInclusion($e_H$)}
\If{$\mathit{edgeIncluded}$}
\If{$\mathit{includedCP} \land$ edgeCount($H,G,e,\overline{\mathit{dir}},y$)$+1 > m_{\mathit{dir}}^{\mathrm{ub}}$}
\State{includeIntoCP($e_H$)}
\State{\Skip}
\EndIf{}
\State{bool $\mathit{swapped} \gets$ randomEdgeSwap($x_G,e^{\prime}_H,\Gsub$)}
\If{!$\mathit{swapped} \land$ edgeCount($H,G,e,\mathit{dir},x$)$+1 > m_{\overline{\mathit{dir}}}^{\mathrm{ub}}$} 
\State{bool $\mathit{resFound} \gets$ resolveBreach($Q,x,e^{\prime}_H,\mathit{dir},e$)} 
\If{$!\mathit{resFound}$}
\State{\Skip} 
\EndIf{}
\EndIf{}
\State{$E_{\Hsub} = E_{\Hsub} \cup \{e^{\prime}_{\Hsub}\}$}
\EndIf{}
\State{$V_{\Hsub} = V_{\Hsub} \cup \{y_{\Hsub}\}$}
\Statex{We might want to include a node $y_{CP}$ even if not required}
\State{bool $\mathit{includedCP} \gets$ randomNodeToCP($Q,x,e^{\prime},\mathit{dir},e$)}
\If{$\mathit{edgeIncluded} \land \mathit{includedCP} \land !\mathit{swapped}$}
\State{randomEdgeToCP()}
\EndIf{}
\EndIf{}
\EndFor{}
\EndFor{}
\EndFor{}
\EndWhile{}
\end{algorithmic}
\end{algorithm}

Algorithm~\ref{alg:constructCP} provides the computations of the split part $\Hsub$ and the crossover point $\CP$. 
We begin with adding all nodes of the border of the problem part to a queue $Q$ (line~2). 
Let $x_H$ be the node we are currently handling (line~3). 
We start by iterating through each of the edges attached to $x_H$ (lines~4--6). 
For each such edge $e_H$, and corresponding neighboring node $y_H$, we first check whether $y_H$ already belongs to $\Hsub$ and, if so, whether the edge $e_H$ also already belongs to $\Hsub$ or $y_H$ already to $\CP$ (lines~7--15). 
If $e_H$ already belongs to $\Hsub$ we can skip the treatment of that edge. 
It means that $y_H$ has already been processed and, hence, $e_H$ adequately been treated. 
If $y_H$ belongs to $\CP$, this makes an important difference for the possible treatment of $e_H$ later on. 
If $y_H$ does not already belong to $\Hsub$ (and, consequently, also not to $\CP$), we can decide randomly if we intend to include it in $\Hsub$ (line~17). 
We then can, again randomly, decide whether we also want to include $e_H$ in $\Hsub$ (line~20). 
Should we decide that we want to include neither $y_H$ nor $e_H$ into $\Hsub$, we can move to the next edge. 

\begin{algorithm}
\caption{\label{alg:randomEdgeSwap}randomEdgeSwap($x,e^{\prime},\Gsub$)}

\begin{algorithmic}[1]
\State{bool $swap \gets$ decideSwap()}
\If{!$swap$}
\State{\Return{\False}}
\Else{}
\State{$e_G \gets$ pickSwapEdgeG($x,e^{\prime},\Gsub$)}
\If{$e_G$ == null}
\State{\Return{\False}}
\Else{}
\State{$E_{\Gsub} \gets E_{\Gsub} \setminus \{e_G\}$}
\State{\Return{\True}}
\EndIf{}
\EndIf{}
\end{algorithmic}
\end{algorithm}

If $y$ already belongs to $\CP$ and, seen in the direction from $y$ to $x$, the upper bound of the type of $e$ has already been reached, there is a single possibility to include $e_H$ in $\Hsub$, namely also including it in $\CP$, that is, identifying it with an edge of the same type between $x$ and $y$ that stems from $\Gsub$. 
Any other way of including $e_H$ in $\Hsub$ would result in an upper bound violation in the to be computed offspring. 
Without providing the details, this procedure is performed by the function \emph{includeIntoCP} (line~23); if inclusion is possible, it is performed, otherwise $e_H$ is not included in $\Hsub$. 
In either case, the remaining iteration for $e_H$ is skipped.

The most options are available if we decide to include both $y_H$ and also $e_H$ in $\Hsub$ (beginning with line~26): 
The first option we have is to swap the edge $e_H$ with another edge (of the same type) adjacent to $x$ that stems from $\Gsub$ (line~26). 
This possibility serves to increase the expressiveness of the operator; recall from Example~\ref{ex:createGsub} that the creation of $\Gsub$ might enforce all edges adjacent to $x_G$ in $G$ to also be part of $\Gsub$. 
Swapping an edge means to choose an edge of the same type as $e_H$ that is attached to $x_G$ in $\Gsub$ and to remove it from $\Gsub$; this procedure is sketched as Algorithm~\ref{alg:randomEdgeSwap}. 
In picking an edge $e_G$ from $\Gsub$ to swap, we have to check whether removing that edge from $\Gsub$ leads to a lower bound violation in the direction opposite to the one in which we are currently working (that check is not detailed in the algorithm). 
If no suitable edge is available, swapping fails. 
Note that this swapping of edges does not affect the multiplicities
of the node $x$ under consideration: 
We remove an edge (that stems from $G$) and replace it by one (that stems from $H$) of the same
type. 

The second option, if we do not decide to swap the edge, is to simply
add it to the offspring (lines~27--31). If we intend to do this, the amount of edges
at the offspring version of $x$ would increase by one. This could
lead to a breach of the maximum. If we detect this, we can try to 
fix the situation by \emph{resolveBreach},  
presented 
\longv{as Algorithm~\ref{alg:resolveBreach} in Appendix~\ref{sec:algorithmic-details}.}
\shortv{in the extended version of our paper~\cite{TK22}.}
Substantially, we need to try adding the pair of $y$ and $e^{\prime}$ to the crossover point.
Doing this would essentially remove the added edge in the offspring,
maintaining our maximum. To add $y$ and $e^{\prime}$ to the crossover point,
we need to find a node $z$ in $\Gsub$, that is of the same type
as $y$, is not part of the crossover point yet, and is connected to
$x$ via an edge of the same type as $e^{\prime}$. 
Furthermore, the edges that are adjacent to that node $z$ in $\Gsub$ may not introduce a violation of an upper bound when combined with the edges that are already adjacent to $y_H$ in $H$ (if such edges exist). 
We search for such a node
by iterating through all edges of the same type as $e^{\prime}$, that are
attached to $x$. If we find such a node, we create a node in the
crossover point, fixing our breach. We then only need to enqueue $y$.
If we cannot find such a node $z$, then we give up on finding a fix
to the breach, instead discarding the addition of $y$ and $e^{\prime}$ to
$\Hsub$. 
For simplicity, we decided to have a failure to resolve the conflict to
result in skipping the current edge type entirely. 
Alternatively, it would be possible to try again to swap the edge.

Once we determined that it is safe to add $y$ and $e^{\prime}$ in the chosen
way, we add them to $\Hsub$ (lines~33 and 35). 
Note that, until now, we have only added $y$ and/or $e^{\prime}$ to the crossover point if we were forced to do so to prevent upper bound violations (via function \emph{includeIntoCP} or \emph{resolveBreach}). 
To increase expressiveness, it should also be possible to randomly include $y$ and, if $y$ is included, possibly also $e^{\prime}$ in $\CP$. 
This functionality is provided by functions \emph{randomNodeToCP} and \emph{randomEdgeToCP} in lines~36--39 of Algorithm~\ref{alg:constructCP}. 
Like in \emph{resolveBreach}, 
adding $y_H$ to the crossover point means choosing a node of the same type
in $\Gsub$ that is not yet included in $\CP$, and adding a new
node to the crossover point, merging them together in the offspring (after performing the necessary checks). 

If we decide to include $y$ in the crossover point, and we did not
swap an edge earlier, we can then choose to also include $e^{\prime}$ in the
crossover point along with $y$. This is only safe to do if we did
not swap any edge. If we decided to both remove an edge in
$\Gsub$, and then combine another edge with $e^{\prime}$, which was set
to replace the deleted edge, we would end up reducing the amount of
edges by one. Since this could breach the minimum, we disallow it
entirely.

\begin{algorithm}
\caption{\label{alg:processFree}processFreeNodes($\Gsub,H,\Hsub$)}

\begin{algorithmic}[1]
\State{queue $Q \gets$ unprocessed nodes from $H$}
\While{$Q$.notEmpty()}
\State{$x_H \gets Q$.pop()}
\State{bool $\mathit{included} \gets x_{H} \in V_{\Hsub}$}
\If{$!\mathit{included}$}
\State{$\mathit{included} \gets$ decideInclusion($x_H$)}
\EndIf{}
\If{$\mathit{included}$}
\State{bool $\mathit{tryCP} \gets$ decideInclusionCP($x_{\Hsub}$)}
\If{$\mathit{tryCP}$}
\State{bool $\mathit{includeCP} \gets$ verifyInclusion($x_{\Hsub}$)}
\EndIf{}
\If{$!\mathit{includeCP}$}
\State{bool $\mathit{minEnsured} \gets$ ensureMinimumFreeNode($x_{\Hsub}$)}
\If{$!\mathit{minEnsured}$}
\State{removeNode($x_{\Hsub}$)}
\Statex{this needs to be cascading the removal as described in the text}
\EndIf{}
\Else{}
\Statex{To find a node to merge $x_{\Hsub}$ with, simply pick a node of equal type in $\Gsub$, that is not yet part of the crossover point}
\State{addToCP($x_{\Hsub}$)}
\State{randomIncludeMoreEdges()}
\Statex{this basically follows $\mathit{ensureMinimumFreeNode}$, just instead of checking if the minimum has been reached, we check whether the maximum has not been reached yet.}
\EndIf{}
\EndIf{}
\EndWhile{}
\end{algorithmic}
\end{algorithm}

Function \emph{constructCP} (Algorithm~\ref{alg:constructCP}) builds a crossover point with all edges
and nodes in $H$ that neighbor a node in $\CP$ being already decided
on. This means that after calling \emph{constructCP}, $\Hsub$ contains all nodes we selected
for the crossover point, along with the nodes and edges chosen to be
included in $\Hsub$ but not $\CP$. 
To further increase expressivity, it should also be possible to include nodes (and their adjacent edges) in $\Hsub$ and possibly $\CP$ that have not been visited so far. 
Furthermore, we might have added nodes to $\Hsub$ without ensuring their lower bound. 
To address both issues, after \emph{constructCP}, \emph{secureCrossover} (Algorithm~\ref{alg:secure-crossover}) calls \emph{processFreeNodes} as introduced in Algorithm~\ref{alg:processFree}.  

We start the function \emph{processFreeNodes} with a queue $Q$ that contains all nodes of $H$ that have not been processed during \emph{constructCP}, i.e., all nodes of $H$ that are not yet part of $\Hsub$ and that never occurred in the queue $Q$ during the call of \emph{constructCP}, and all nodes that belong to $\Hsub$ without also belonging to $\CP$. 
These are nodes for which we never decided whether or not they should be part of $\Hsub$ and $\CP$, or for which we did not yet ensure satisfaction of lower bounds. 
In \emph{processFreeNodes}, we first check whether or not the currently considered node $x_H$ is already part of $\Hsub$ and, if not, give the node a random chance of being included in $\Hsub$ (line~4--7). 
(While the check is unnecessary at the very start of the algorithm, it is necessary later on when nodes from $Q$ that have not been reached yet might have already been included in $\Hsub$, triggered by the inclusion of adjacent edges.) 
In the next step, if the node is to be included, we randomly decide whether it also shall be included in $\CP$ (line~9); the function \emph{decideInclusionCP} is understood to (i) randomly decide whether or not $x_H$ shall be included in $\CP$ and, if so, to check whether or not a node of the same type that does not yet belong to $\CP$ is available in $\Gsub$. 
This leaves us with two situations: 
If $x_H$ is to be included in $\CP$, it gets identified with a node from $\Gsub$ that brings enough edges to satisfy lower bounds (provided feasibility of $G$). 
However, we need to ensure that the identification does not introduce violations of upper bounds (in case $\Hsub$ already contains edges adjacent to $x_H$). 
If $x_H$ is to be included in $\Hsub$ but not in $\CP$, it needs to be accompanied by enough edges to not introduce violations of lower bounds. 
Inclusion of these edges in $\Hsub$, however, might not be possible without introducing violations of upper bounds at their respective other adjacent nodes (if those are part of $\CP$). 
These are the purposes of functions \emph{verifyInclusion} and \emph{ensureMinimumFreeNode}, called in lines~11 and 14 of Algorithm~\ref{alg:processFree}, respectively. 

Function \emph{verifyInclusion} 
verifies that the inclusion of $x_H$ in $\CP$ does not breach any maxima in the offspring. 
We do this by performing the same check as in \emph{constructCP} (Algorithm~\ref{alg:constructCP}), however, 
we stop at merely checking whether a breach would occur. 
Of course, one could also try to find a fix for potentially arising breaches of a maximum, via edge swapping, or recursive inclusion of further nodes, along with the edges. 
For simplicity, and since the logic is already shown prior, we omitted this.
The basic structure of \emph{verifyInclusion} is given 
\longv{as Algorithm~\ref{alg:verifyInclusion} in Appendix~\ref{sec:algorithmic-details}.}
\shortv{in the extended version of this paper~\cite{TK22}.} 

\begin{algorithm}
\caption{\label{alg:ensureMinFreeNode}ensureMinimumFreeNode($x$)}

\begin{algorithmic}[1]
\ForAll{$\mathit{dir} \in \{\mathit{src},\mathit{tar}\}$}
\ForAll{$e_{\TG} \in E_{\TG}$ with $\mathit{dir}_{\TG}(e_{\TG})=x_{\TG}$}
\State{int $\mathit{targetAmount} \gets \min{(|\text{edges(}H,e_{\TG},\mathit{dir},x_{H}\text{)}|, m_{\overline{\mathit{dir}}}^{\mathrm{lb}})}$}
\ForAll{$e^{\prime} \in$ edges($H,e_{\TG},\mathit{dir},x_{H}$)}
\If{edgeCount($H,G,e_{\TG},\overline{\mathit{dir}},\overline{\mathit{dir}}(e^{\prime})$)$+1 \leq m_{\mathit{dir}}^{\mathrm{ub}}$}
\State{$E_{\Hsub} \gets E_{\Hsub} \cup \{e^{\prime}\}$}
\If{!$\overline{\mathit{dir}}(e^{\prime}) \in \Hsub$}
	\State{$V_{\Hsub} \gets V_{\Hsub} \cup \{\overline{\mathit{dir}}(e^{\prime})\}$}
	\State{$Q$.enqueue($\overline{\mathit{dir}}(e^{\prime})$)}
\EndIf{}
\EndIf{}
\If{$|$edges($\Hsub,e_{\TG},\mathit{dir},x_{\Hsub}$)$| \geq \mathit{targetAmount}$}
\State{\Return{\True}}
\EndIf{}
\EndFor{}
\EndFor{}
\EndFor{}
\State{\Return{\False}}
\end{algorithmic}
\end{algorithm}

For the nodes that did not get picked to be included in the crossover point,
or that were rejected during verification, we try to ensure that
we include enough edges to meet the minimum via \emph{ensureMinimumFreeNode} (Algorithm~\ref{alg:ensureMinFreeNode}). 
During the inclusion, we ensure that the inclusion of a selected edge does not introduce a violation of an upper bound in the opposite direction (line~6). 
If the second adjacent node of an included edge does not already belong to $\Hsub$, we add it and newly enqueue it to ensure that also the lower bounds of that node are controlled. 
We use $\mathit{targetAmount}$ to only allow the lower bound to be violated if $H$ already violated the lower bound in this position. Like with $G$, this allows us to meaningfully treat the case where the input is infeasible: We ensure that the amount of edges in the offspring is equal to that in the input, meaning that secure crossover cannot worsen the situation by introducing a greater number of missing edges. 

Returning to \emph{processFreeNodes} (Algorithm~\ref{alg:processFree}), should it prove impossible to include a node in $\Hsub$ while ensuring its minimum (i.e., \emph{ensureMinimumFreeNode} returns false), the easiest solution is to
simply remove the node that failed to reach the minimum from $\Hsub$,
along with all edges. This removal has the potential to cascade, meaning
other free nodes might have relied on a removed edge for their own
minimum. We remove all nodes affected by this cascading effect. More
elaborate solutions that try to remedy each newly affected node are
imaginable, but in the interest of simplicity we chose a more coarse
approach.

All nodes we added to reach minima for our original free nodes are
added to the queue of free nodes, since we will need to ensure their
feasibility as well. Following this process, we iterate through $H$ in its entirety. The iteration stops either when all minima are reached and all nodes had a chance to be included, or when the cascading determines that no nodes aside the ones in $\CP$ are possible additions to $\Hsub$, since there is always a failure to meet minima. 

Our approach therefore visits each node in $G$ exactly once (when constructing $\Gsub$) and the nodes in $H$ are visited at most thrice, once or twice for constructing $\Hsub$ and once when potentially removing nodes for failure to meet minima. This, along with the fact that we only perform checks against the multiplicities when we cannot rely on feasibility inherited by the input, leads us to believe that using secure crossover is beneficial compared to simpler approaches, e.g., using generic crossover and discarding infeasible offsprings. 

\begin{example}
	In Example~\ref{ex:example-generic-crossover}, we discussed how generic crossover can compute the solution $\overline{G_1H_2}$, depicted in Fig.~\ref{fig:example-offspring}, from the solutions $\overline{G}$ and $\overline{H}$, depicted in Fig.~\ref{fig:example-solutions}. 
	Secure crossover as introduced in this section can also compute this solution from the same input. 
	As stated in Example~\ref{ex:createGsub}, \emph{createGsub} (Algorithm~\ref{alg:sichererSplit-1}) will first return $G$ as $\Gsub$. 
	However, \emph{constructCP} (Algorithm~\ref{alg:constructCP}) then iterates over the three \textsf{Classes c2, c3}, and \textsf{c4} of $\overline{H}$ because these constitute the border in that case. 
	Since none of these \textsf{Classes} already belongs to $\Hsub$, for each of it it is randomly decided whether or not it (and its adjacent edge) are to be included in $\Hsub$. 
	Assume that a run of \emph{constructCP} randomly decides (i) to include \textsf{c2} (and its adjacent edge) in $\Hsub$ and $\CP$, (ii) to include \textsf{c4} only in $\Hsub$ and not in $\CP$ and to swap its adjacent edge against the one from $\Gsub$ pointing to \textsf{Feature f3} (which means to delete that edge from $\Gsub$), and (iii) to not include \textsf{c3} in $\Hsub$. 
	Then $\Gsub$ coincides with $G_1$ from Example~\ref{ex:example-generic-crossover}, $\Hsub$ with $H_2$, and also the crossover points coincides. 
	This means, the relevant offspring computed by secure crossover in this case is the feasible solution $\overline{G_1H_2}$.
	
	Note, first, that, while generic crossover can of course also compute $\overline{G_1H_2}$, the probability that it ever returns feasible solutions drastically decreases with increasing size and complexity of the input graphs. 
	Second, because of the structural simplicity of the CRA case, the call of function \emph{processFreeNodes} (Algorithm~\ref{alg:processFree}) is trivial in that case. 
	It is initialized with the empty queue because all nodes of $H$ have already been processed by \emph{constructCP}.
\end{example}

%% file: formal-properties.tex
In the following, we present the central formal results for our crossover operator. 
First, we prove that offspring computed from feasible solutions is again feasible, and, secondly, that every feasible solution is a possible result of the application of secure crossover or, in other words, that, in principle, we only restrict the reachable search space by infeasible solutions. 
Both of these properties are valuable since we want to obtain a crossover
operator that guarantees that the output models are feasible
without restricting these output models in any way not required by
the feasibility criteria. 
\longv{In Appendix~\ref{sec:proofs},} 
\shortv{In the extended version of this paper~\cite{TK22},} 
we additionally establish that the application of secure crossover terminates and that it is an instance of the generic approach suggested in~\cite{TJK22}. 
In particular, this ensures that it computes solutions for the given problem instance. 

In all of the following, we always assume two solutions $\overline{G}$ and $\overline{H}$ for the same problem instance $\overline{\PI}$ to be given. 
As already explained, with secure crossover we are actually only interested in the first computed offspring, namely the one that arises as union of $\Gsub$ and $\Hsub$ over the chosen crossover point $\CP$. 
We just denote that offspring as $O$ and ignore the second computed offspring. 

\begin{theorem}[Feasibility of offspring models]
\label{prop:Korrektheit}

Let $\overline{G}, \overline{H}$ be two solutions of a problem instance and let $\overline{O}$ be an
offspring that has been computed by secure crossover. It holds that:

\begin{enumerate}
	\item For every node $x_{O}$ in $O$ the number of edges $e^{\prime}_{O}$ of a type $e_{TG}$ with $\mathit{dir}(e^{\prime}_{O})=x_{O}$ is greater or equal to the number of edges minimally provided by its counterparts in $G$ and/or $H$, that is, greater or equal to
	\begin{equation*}
		\min\bigg( \Big\{ m_{\overline{\mathit{dir}}}^{\mathrm{lb}}(e_{TG}),
							 \left|\left\{ e^{\prime}_{G}\in E_{G} \mid t_G(e^{\prime}_{G})=e_{TG},\mathit{dir}(e^{\prime}_{G})=x_{G}\right\} \right| \Big\} \bigg), 
	\end{equation*}
	or an analogous minimum should $x_O$ only have a counterpart in $H$. 
	
	\item For every node $x_{O}$ in $O$ the number of edges $e^{\prime}_{O}$ of a type $e_{TG}$ with $\mathit{dir}(e^{\prime}_{O})=x_{O}$ is smaller or equal to the number of edges maximally provided by its counterparts in $G$ and/or $H$, that is, smaller or equal to
	\begin{equation*}
		\max\bigg(	\Big\{ m_{\overline{\mathit{dir}}}^{\mathrm{ub}}(e_{TG}),
								\left|\left\{ e^{\prime}_{G}\in E_{G} \mid t_G(e^{\prime}_{G})=e_{TG},\mathit{dir}(e^{\prime}_{G})=x_{G}\right\} \right| \Big\} \bigg), 
	\end{equation*}
	or an analogous maximum should $x_O$ only have a counterpart in $H$. 
\end{enumerate}
In particular, if $\overline{G}$ and $\overline{H}$ are feasible, so is $\overline{O}$.
\end{theorem}

The proof of the above theorem is split in two parts, which we build from multiple lemmas.
The basic idea is that we separately show that the offspring models
satisfy all the lower bounds of the multiplicities and that the offspring
models satisfy the upper bounds. For this we will first prove that
our selection of $\Gsub$ always provides a feasible model, provided feasibility of $G$.

\begin{lemma}[Construction of $\Gsub$]
\label{lem:gsubconstruction}
 
Let $\Gsub$ be a subgraph of $G$ that has been obtained by a call of \emph{createGsub} (Algorithm~\ref{alg:sichererSplit-1}). 
Then, for every node $x_{\Gsub}$ in $\Gsub$ the number of edges $e^{\prime}_{\Gsub}$
of a type $e_{TG}$ with $\mathit{dir}(e^{\prime}_{\Gsub})=x_{\Gsub}$ is greater
or equal to 
\begin{equation*}
\min\bigg(\Big\{ m_{\overline{\mathit{dir}}}^{\mathrm{lb}}(e_{TG}),
					\left|\left\{ e^{\prime}_{G}\in E_{G} \mid t_G(e^{\prime}_{G})=e_{TG},\mathit{dir}(e^{\prime}_{G})=x_{G}\right\} \right| \Big\} \bigg). 
	\end{equation*}
\end{lemma}

\begin{lemma}[Offspring satisfies lower bounds at least as well as input]
\label{lem:offspring-meets-lbs}
Let $\overline{G}, \overline{H}$ be two solutions of a problem instance and let $\overline{O}$ be an
offspring that has been computed by secure crossover. 
Let $\TG_{\mathrm{mod}}$ be a modified type graph, where each edge multiplicity $[i,j]$ has been replaced with $[k,*]$, where $k$ denotes the minimum of $i$ and the minimal number of edges of that type and direction that appear adjacent to a node in $G$ or $H$ (i.e., the worst violation of the lower bound). 
Then $O$ is feasible with
regard to all edge multiplicities in $\TG_{\mathrm{mod}}$, and for every node
$x_{O}$ in $O$ the number of edges $e^{\prime}_{O}$ of one type $e_{TG}$
with $\mathit{dir}(e^{\prime}_{O})=x_{O}$ is greater or equal to 
\begin{equation*}
\min\bigg(\Big\{ m_{\overline{\mathit{dir}}}^{\mathrm{lb}}(e_{TG}),
					\left|\left\{ e^{\prime}_{G}\in E_{G} \mid t_G(e^{\prime}_{G})=e_{TG},\mathit{dir}(e^{\prime}_{G})=x_{G}\right\} \right| \Big\} \bigg). 
	\end{equation*}
or an analogous minimum for $H$, should $x_O$ only stem from there. 
In particular, if $G$ and $H$ are feasible, $O$ is feasible for the type graph $\TG_{\mathrm{min}}$, where each edge multiplicity $[i,j]$ has been replaced with $[i,*]$.
\end{lemma}

We now show that the offspring satisfies the upper bounds, too.
\begin{lemma}[Offspring satisfies upper bounds]
\label{lem:satisfaction-upper-bounds}

Let $\overline{G}, \overline{H}$ be two solutions of a problem instance and let $\overline{O}$ be an
offspring that has been computed by secure crossover.
Let $\TG_{\mathrm{mod}}$ be a modified type graph, where each edge multiplicity $[i,j]$ has been replaced with $[0,k]$, where $k$ denotes the maximum of $j$ and the maximal number of edges of that type and direction that appear adjacent to a node in $G$ (i.e., the worst violation of the upper bound in $G$). 
Then $O$ is feasible with
regard to all edge multiplicities in $\TG_{\mathrm{mod}}$, and for every node
$x_{O}$ in $O$ the number of edges $e^{\prime}_{O}$ of one type $e_{TG}$
with $\mathit{dir}(e^{\prime}_{O})=x_{O}$ is smaller or equal to 
\begin{equation*}
		\max\bigg(	\Big\{ m_{\overline{\mathit{dir}}}^{\mathrm{ub}}(e_{TG}),
								\left|\left\{ e^{\prime}_{G}\in E_{G} \mid t_G(e^{\prime}_{G})=e_{TG},\mathit{dir}(e^{\prime}_{G})=x_{G}\right\} \right| \Big\} \bigg). 
	\end{equation*}

In particular, if $G$ and $H$ are feasible, $O$ is feasible for the type graph $\TG_{\mathrm{max}}$, where each edge multiplicity $[i,j]$ has been replaced with $[0,j]$.
\end{lemma}

\begin{proposition}[Coverage of search space]
\label{prop:coverage-search-space}

Let $O$ be a feasible solution graph of the problem instance $\PI$.
Then there exist two feasible solution graphs $G$ and $H$ of $\PI$, different from $O$, such that $O$ is an offspring model of a secure
crossover of $G$ and $H$.
\end{proposition}

%% file: conclusion.tex
In this paper, we develop \emph{secure crossover}, a crossover operator on typed graphs (or, more generally, computation graphs) that preserves multiplicity constraints. 
This means, applied to feasible input, the secure crossover computes feasible output. 
Even for infeasible input, an application of secure crossover at least does not worsen the amount of multiplicity violations. 
Secure crossover is an extensive refinement of \emph{generic crossover} as introduced in~\cite{TJK22}; to simplify the computation of secure crossover, we focus on the computation of a single (feasible) offspring solution. 
We develop the underlying algorithms of secure crossover in great detail, and use this to prove central formal properties of it, namely the preservation of feasibility, that all feasible solutions can, in principle, still be computed as offspring of an application of secure crossover, and that secure crossover is an instance of generic crossover. 
Beyond its immediate applicability in MDO, our work is a further indication that MDO is a promising approach when one is interested in guaranteeing certain properties of search operators during (meta-heuristic) search: we are able to verify a highly complex property (preservation of multiplicity constraints) for a still quite generic crossover operator on graphs.  
It seems at least intuitive that properties of a similar complexity are hardly verifiable when working, e.g., on Bit-strings as a representation for solutions during search. 

With regards to future work, we intend to implement secure crossover and to investigate whether evolutionary search on models can profit from the preservation of feasibility. 
With its comprehensive formal basis at hand, it would be interesting to even provide a verified implementation of secure crossover. 
We also intend to extend our construction to take further kinds of constraints into account, e.g., nested graphs constraints (which are equivalent to first-order logic on graphs) as, for example, presented in~\cite{HP09}.

%% file: algorithmic-details.tex
We first provide the details for the inclusion of edges in $\Gsub$ (compare Sect.~\ref{sec:creation-Gsub}).
These details are presented in Algorithm~\ref{alg:ensureMinimum-1}. 
Given a node $x$ for which adjacent edges are to be included, for each type of edge for which $x$ can either serve as a source or a target node (lines~1--2), we determine a random number $n$ of edges that are to be included (line~3). 
This random number is selected from a specific range: 
The function $\mathit{lb}(e_{\TG})$ returns the minimum of $\lvert edges(G,e_{\TG},dir,x_G)\rvert$, which gives the number of actually adjacent edges of the considered type to $x_G$ in $G$, and $m_{\overline{dir}}^{\mathrm{lb}}(e_{\TG})$, i.e., the required lower bound; moreover, $\overline{\mathit{src}} = \mathit{tar}$ and $\overline{\mathit{tar}} = \mathit{src}$.
Similarly, $\mathit{ub}(e_{\TG})$ returns the \emph{minimum} of $\lvert edges(G,e_{\TG},dir,x_G)\rvert$ and $m_{\overline{dir}}^{\mathrm{ub}}(e_{\TG})$, i.e., the required upper bound. 
In this way, $n$ lies between the lower and the upper bound of the considered edge type, or it coincides with the number of available edges of that type (at that node), should the number of edges available in $G$ fall short of the lower bound. 
Then, $n$ edges of the considered type, connected to the node $x$ in the considered direction are randomly selected to be included in $\Gsub$. 
This inclusion comprises the inclusion of their second attached node. 
Without representing this explicitly in the code, we assume (i) that the number $n$ of edges that are to be included is reduced by the number of edges that already have been included so far (when including adjacent edges for another node) and (ii) that it is ensured that for newly included nodes the function \emph{includeAdjacentEdges} will be called in the future.

\begin{algorithm}
\caption{\label{alg:ensureMinimum-1}includeAdjacentEdges($x,G$)}

\begin{algorithmic}[1]
\ForAll{$dir \in \{\mathit{src}_G,\mathit{tar}_G\}$}
\ForAll{$e_{\TG} \in E_{\TG}$ with $dir_{\TG}(e_{\TG})=t_G(x)$}
\State{$n \gets$ randInt($\mathit{lb}(e_{\TG}), \mathit{ub}(e_{\TG})$)}
\For{$i\gets 1, \dots, n$}
\State{select $e^{\prime} \in \mathit{edges}(G,e_{\TG},dir,x_{G})$}
\State{$E_{\Gsub} \gets E_{\Gsub} \cup \{e^{\prime}\}$}
\State{$V_{\Gsub} \gets V_{\Gsub} \cup \{\overline{dir}_G(e^{\prime})\}$}
\EndFor{}
\EndFor{}
\EndFor{}
\end{algorithmic}
\end{algorithm}

Next, we provide the pseudocode for function \emph{resolveBreach} (Algorithm~\ref{alg:resolveBreach}) that can be called during the computation of $\Hsub$ and $\CP$.

\begin{algorithm}
\caption{\label{alg:resolveBreach}resolveBreach($Q,x,e^{\prime},\mathit{dir},e_{\TG}$)}

\begin{algorithmic}[1]
\State{$y_G \gets$ pickFreeNodeG($x_G,\mathit{dir},e_{\TG}$)}
\If{$y_G == null$}
\State{\Return{\False}}
\EndIf{}
\State{$V_{CP} = V_{CP} \cup \{y_{CP}\}$}
\State{$E_{CP} = E_{CP} \cup \{e^{\prime}_{CP}\}$}
\State{$Q$.enqueue($y_H$)}
\State{\Return{\True}}
\end{algorithmic}
\end{algorithm}

Finally, Algorithm~\ref{alg:verifyInclusion} provides the basic structure for the function \emph{verifyInclusion} that is used to check whether it is possible to include a node in $\CP$.

\begin{algorithm}
\caption{\label{alg:verifyInclusion}verifyInclusion($x$)}

\begin{algorithmic}[1]
\ForAll{$\mathit{dir} \in \{\mathit{src},\mathit{tar}\}$}
\ForAll{$e_{\TG} \in E_{\TG}$ with $\mathit{dir}_{\TG}(e_{\TG})=t_H(x)$} 
\State{$over \gets$ edgeCount($\Gsub,\Hsub,e_{\TG},\mathit{dir},x$) $- \lvert$edges($\CP,e_{\TG},\mathit{dir},x_{\CP}$)$\rvert > max_{\overline{\mathit{dir}}}$}
\If{$over$}
\Statex{Here we could implement varying degrees of fixing attempts. From edge swapping, to recursive merging}
\State{\Return{\False}}
\EndIf{}
\EndFor{}
\EndFor{}
\State{\Return{\True}}
\end{algorithmic}
\end{algorithm}

%% file: proofs.tex
We first state that every call of secure crossover terminates.

\begin{proposition}[Termination of \emph{secure crossover}]
	\label{prop:termination-secure-crossover}
	
	Given a computation type graph with multiplicities $\overline{\TG}$, a problem instance $\overline{\PI}$ for it, and two solutions $\overline{G}$ and $\overline{H}$ for $\overline{\PI}$, every application of \emph{secure crossover} as defined in Algorithm~\ref{alg:secure-crossover} terminates, computing two offspring graphs $\overline{O_1}$ and $\overline{O_2}$. 
\end{proposition}

\begin{proof}
	First, none of the algorithms that are called during the application of secure crossover throws an exception; at most, certain iterations of a loop are skipped (e.g., in Algorithm~\ref{alg:constructCP}) or a restricted set of computations is rolled back (e.g., in Algorithm~\ref{alg:processFree}). 
	This means that, if secure crossover terminates, it definitely stops with the computation of the two offspring graphs $\overline{O_1}$ and $\overline{O_2}$.
	
	The only points that could cause non-termination are the while loops in Algorithms~\ref{alg:constructCP}, \ref{alg:processFree}, and \ref{alg:ensureMinFreeNode}. 
	Both Algorithms~\ref{alg:constructCP} and \ref{alg:processFree} iterate over a queue $Q$ that is initialized with a finite number of nodes. 
	In both cases, additional nodes can be enqueued during the computation. 
	However, in both cases this only happens for nodes that are newly included in $\CP$ (Algorithm~\ref{alg:resolveBreach}, line~7) or newly included in $\Hsub$ (Algorithm~\ref{alg:ensureMinFreeNode}, line~9). 
	For every node of $H$, this can only happen once, and in all cases, it is first checked whether or not the inclusion already holds. 
	In Algorithm~\ref{alg:ensureMinFreeNode}, both while-loops are guarded by counters that in- or decrease with every iteration. 
	
	Overall, secure crossover terminates. 
\end{proof}

As recalled in Sect.~\ref{sec:preliminaries}, applying the generic crossover operator from~\cite{TJK22} to computation graphs $\overline{G}$ and $\overline{H}$ amounts to splitting these into computation graphs $\overline{G_1}, \overline{G_2}, \overline{H_1}$, and $\overline{H_2}$, each containing the considered problem graph, and uniting $G_1$ and $H_2$ and $G_2$ and $H_1$, respectively, over a common crossover point $\overline{\CP}$ that is a common subgraph of the split points $\GI$ and $\HI$ and also at least contains the considered problem graph. 
Proposition~2 in~\cite{TJK22} clarifies that the computed offspring are again computation graphs and, in particular, solutions for the considered problem instance $\PI$. 
While it is also straightforward to directly prove this claim for our crossover operator, we just obtain it as a corollary to the general result from~\cite{TJK22} because our crossover operator is just a special implementation of that procedure.

\begin{lemma}[Secure crossover as generic crossover]
	\label{lem:secCrossAsInstance}
	Given a computation type graph with multiplicities $\overline{\TG}$, a problem instance $\overline{\PI}$ for it, and two solutions $\overline{G}$ and $\overline{H}$ for $\overline{\PI}$, applying \emph{secure crossover} as defined in Algorithm~\ref{alg:secure-crossover} is a special instance of applying generic crossover as defined in~\cite{TJK22}.
\end{lemma}

\begin{proof}
	By construction, $\Gsub$ and $\Hsub$, which take that parts of $G_1$ and $H_2$ in generic crossover, are subgraphs of $G$ and $H$, respectively, that contain the given problem graph $\PIP$; also the constructed crossover point $\CP$ always contains $\PIP$. 
	As $G_2$ and $H_1$ we choose $G$ and $H$, respectively, which results in the split points being given as $\GI = \Gsub$ and $\HI = \Hsub$. 
	Finally, $\CP$ is a common subgraph of $\Gsub$ and $\Hsub$. 
	Hence, the construction of secure crossover implements a specific variant of generic crossover.
\end{proof}

\begin{corollary}[Structural correctness of offspring]
	\label{cor:structural-correctness}
	Given a computation type graph with multiplicities $\overline{\TG}$, a problem instance $\overline{\PI}$ for it, and two solutions $\overline{G}$ and $\overline{H}$ for $\overline{\PI}$, applying \emph{secure crossover} as defined in Algorithm~\ref{alg:secure-crossover} results in two computation models $\overline{O_1}$ and $\overline{O_2}$ that are solutions for $\overline{\PI}$.
\end{corollary}

\begin{proof}
	Because of Lemma~\ref{lem:secCrossAsInstance} this is just a corollary to \cite[Proposition~2]{TJK22}.
\end{proof}

In the rest of this section, we prove the results of the main paper.

\begin{proof}[Proof of Lemma~\ref{lem:gsubconstruction}]
Let $G$ be a solution model as described in the lemma. Let $x_{\Gsub}$
be a node from $V_{\Gsub}$, $e_{TG}$ an edge type and $\mathit{dir}$ a
direction. The function \emph{createGsub} iterates through every node
in $V_{G}$, so eventually, it will consider $x_{G}$, the equivalent
node for $x_{\Gsub}$ in $G$. For $x_{G}$, we first check whether
it has already been included in $\Gsub$. Since we have our node
$x_{\Gsub}$, we can assume it is included. This means that in line~8
we call the function \emph{includeAdjacentEdges} (Algorithm~\ref{alg:ensureMinimum-1}) for $x_{G}$.

In this function, we iterate through every direction, and every type
that an edge connected to $x_{G}$ can have, so we will inevitably
also run the loop\rq{}s body for $e_{TG}$ and $\mathit{dir}$. 
In that case, we include a randomly selected number $n$ of edges (of the according type and direction), ensuring that the relevant lower bound is reached, or at least all possible edges are included (line~3). 
\end{proof}

\begin{proof}[Proof of Lemma~\ref{lem:offspring-meets-lbs}]
Let $G,H$ be the input solutions and $O$ be the computed offspring. 
Let $e_{\TG_{\mathrm{mod}}}\in E_{\TG_{\mathrm{mod}}}$ be an edge from the modified
type graph, $\mathit{dir}$ one of the possible direction $\mathit{src}$ or $\mathit{tar}$ for the
edges, and 
\begin{equation*}
p_{O}\in\left\{ x\in V_{O} \mid t_O(x)=\mathit{dir}(e_{\TG_{\mathrm{mod}}})\right\} 
\end{equation*}
a node from the offspring that can have edges of type $e_{\TG_{\mathrm{mod}}}$.
For a set of edges to satisfy $m_{\overline{\mathit{dir}}}(e_{\TG_{\mathrm{mod}}})$, we can
see that it suffices that 
\begin{equation*}
\left|\left\{ e^{\prime}_{O}\in E_{O} \mid t_O(e^{\prime}_{O})=e_{\TG_{\mathrm{mod}}},\mathit{dir}(e^{\prime}_{O})=p_{O}\right\} \right|\geq k,
\end{equation*}
where $k$ is the new, relaxed lower bound of $\TG_{\mathrm{mod}}$, 
since the second part of the relaxed multiplicity is always satisfied (as $j=*$).
We might therefore write 
\begin{equation*}
\left|\left\{ e^{\prime}_{O}\in E_{O} \mid t_O(e^{\prime}_{O})=e_{\TG_{\mathrm{mod}}},\mathit{dir}(e^{\prime}_{O})=p_{O}\right\} \right|\geq k
\end{equation*}
to note that the entire multiplicity is satisfied. %

The node $p_{O}$ can stem either solely from $\Gsub$, with no
 preimage in $\CP$, solely from $\Hsub$, or from both, with
a preimage $p_{\CP}$ in $V_{\CP}$. We therefore observe the
three cases:
\begin{enumerate}
\item $p_{O}$ stems solely from $\Gsub$:\\
In this case, all edges $e^{\prime}_{O}$ with $t_O(e^{\prime}_{O})=e_{TG}$ and $\mathit{dir}(e^{\prime}_{O})=p_{O}$
must also stem solely from $\Gsub$, without a preimage in
$\CP$. This is due to the fact that $\CP\hookrightarrow \Gsub$ is
a graph morphism. 
This means that 
\begin{equation*}
	\begin{split}
		\left|\left\{ e^{\prime}_{O}\in E_{O} \mid t_O(e^{\prime}_{O})=e_{\TG_{\mathrm{mod}}},\mathit{dir}(e^{\prime}_{O})=p_{O}\right\} \right|=\\
		\left|\left\{ e^{\prime}_{\Gsub}\in E_{\Gsub} \mid t_{\Gsub}(e^{\prime}_{\Gsub})=e_{\TG_{\mathrm{mod}}},\mathit{dir}(e^{\prime}_{\Gsub})=p_{\Gsub}\right\} \right|,
	\end{split}
\end{equation*}
and Lemma~\ref{lem:gsubconstruction} then proves the statement. 
\item $p_{O}$ stems solely from $\Hsub$:\\
Again, this means all edges $e^{\prime}_{O}$ with $t_O(e^{\prime}_{O})=e_{TG}$ and
$\mathit{dir}(e^{\prime}_{O})=p_{O}$ must also stem solely from $\Hsub$. During
the construction of $\Hsub$, each node that we include must be processed by Algorithm~\ref{alg:processFree}. 
This means that in line~14, we called function \emph{ensureMinimumFreeNode} for that respective node. 
Since $p_{O}$ is part of $\Hsub$, that call returned true, meaning that the node either satisfies the respective lower bound or is at least attached to as many edges as are provided by $H$. 

\item There exists a preimage $p_{\CP}$ for $p_{O}$:\\
To begin with, if $p_{O}$ has a preimage $p_{\CP}$ in the crossover point,
then there must be a $p_{\Gsub}$ in $\Gsub$. The number of edges
at $p_{O}$ must be therefore be greater or equal than the number
of edges at $p_{\Gsub}$ at the start of the construction of the
$\CP$. As seen in Case (1) this number of edges satisfies the requirement.
During construction of the crossover point, we only reduce the number
of edges in one instance, that is when we swap edges. If we swap edges (Algorithm~\ref{alg:randomEdgeSwap}), 
we always include a new edge to replace the one we removed. We also
then ensure in line~37 of Algorithm~\ref{alg:constructCP} that our newly included replacement does not
get included in $\CP$, thus maintaining the same number of edges.
All other instances merely add edges.
\end{enumerate}

The statement for the case of feasible input is an immediate consequence of the above considerations because, in that case, the original lower bound $i$ coincides with the relaxed lower bound $k$.
\end{proof}

\begin{proof}[Proof of Lemma~\ref{lem:satisfaction-upper-bounds}]
Let $G,H$ be the input solutions and $O$ be the computed offspring. 
Let $e_{\TG_{\mathrm{mod}}}\in E_{\TG_{\mathrm{mod}}}$ be an edge from the modified
type graph, $\mathit{dir}$ one of the possible direction $\mathit{src}$ or $\mathit{tar}$ for the
edges, and 
\begin{equation*}
p_{O}\in\left\{ x\in V_{O} \mid t_{TG_{\mathrm{mod}}}(x)=\mathit{dir}(e_{\TG_{\mathrm{mod}}})\right\} 
\end{equation*}
a node from the offspring that can have edges of type $e_{\TG_{\mathrm{mod}}}$.
For a set of edges to satisfy $m_{\overline{\mathit{dir}}}(e_{\TG_{\mathrm{mod}}})$, we can
see that it suffices that 
\begin{equation*}
\left|\left\{ e^{\prime}_{O}\in E_{O} \mid t_O(e^{\prime}_{O})=e_{\TG_{\mathrm{mod}}},\mathit{dir}(e^{\prime}_{O})=p_{O}\right\} \right|\leq k,
\end{equation*},
where $k$ is the relaxed upper bound of $\TG_{\mathrm{mod}}$, 
since the first part of the multiplicity is always true (as $i=0$).
We might therefore write 
\begin{equation*}
\left|\left\{ e^{\prime}_{O}\in E_{O} \mid t_O(e^{\prime}_{O})=e_{\TG_{\mathrm{mod}}},\mathit{dir}(e^{\prime}_{O})=p_{O}\right\} \right|\leq k
\end{equation*}
to note that the entire multiplicity is satisfied.

The node $p_{O}$ can stem either solely from $\Gsub$, with no
inverse image in $\CP$, solely from $\Hsub$, or from both, with
an inverse image $p_{\CP}$ in $V_{\CP}$. We therefore observe the
three cases:
\begin{enumerate}
\item $p_{O}$ stems solely from $\Gsub$:\\
In this case, all edges $e^{\prime}_{O}$ with $t_O(e^{\prime}_{O})=e_{TG}$ and $\mathit{dir}(e^{\prime}_{O})=p_{O}$
must also stem solely from $\Gsub$, without an inverse image in
$\CP$. This is due to the fact that $\CP\hookrightarrow \Gsub$ is
a graph morphism. This means that 
\begin{equation*}
	\begin{split}
		\left|\left\{ e^{\prime}_{O}\in E_{O} \mid t_O(e^{\prime}_{O})=e_{\TG_{\mathrm{mod}}},\mathit{dir}(e^{\prime}_{O})=p_{O}\right\} \right|= \\
		\left|\left\{ e^{\prime}_{\Gsub}\in E_{\Gsub} \mid t_{\Gsub}(e^{\prime}_{\Gsub})=e_{\TG_{\mathrm{mod}}},\mathit{dir}(e^{\prime}_{\Gsub})=p_{\Gsub}\right\} \right|,
	\end{split}
\end{equation*}
which is smaller than $k$ by the definition of $k$. 

\item $p_{O}$ stems solely from $\Hsub$:\\
Again, this means all edges $e^{\prime}_{O}$ with $t_O(e^{\prime}_{O})=e_{TG}$ and
$\mathit{dir}(e^{\prime}_{O})=p_{O}$ must also stem solely from $\Hsub$. Since $\Hsub$
is a subgraph of $H$, it follows that 
\begin{equation*}
	\begin{split}
		\left|\left\{ e^{\prime}_{O}\in E_{O} \mid t_O(e^{\prime}_{O})=e_{\TG_{\mathrm{mod}}},\mathit{dir}(e^{\prime}_{O})=p_{O}\right\} \right|= \\
		\left|\left\{ e^{\prime}_{\Hsub}\in E_{\Hsub} \mid t_{\Hsub}(e^{\prime}_{\Hsub})=e_{\TG_{\mathrm{mod}}},\mathit{dir}(e^{\prime}_{\Hsub})=p_{\Hsub}\right\} \right| .
	\end{split}
\end{equation*}
As every inclusion of an edge into $\Hsub$ is checked for not introducing an upper bound violation (in any direction), in this case we can even conclude that the original upper bound $j$ is satisfied. 

\item There exists an inverse image $p_{\CP}$ for $p_{O}$\\
To begin with, if $p_{O}$ has an inverse image $p_{\CP}$ in the crossover point,
then there must be a $p_{\Gsub}$ in $\Gsub$. The number of edges
at $p_{O}$ must be therefore be equal to the number of edges at $p_{\Gsub}$
at the start of the construction of the $\CP$. As seen in Case (1),
this number of edges satisfies the relaxed multiplicity $k$. During construction
of the crossover point, we only increase the number of edges after
specifically checking that the inclusion does not breach the maximum.
\end{enumerate}

Again,the statement for the case of feasible input is an immediate consequence of the above considerations because, in that case, the original upper bound $j$ coincides with the relaxed upper bound $k$.
\end{proof}

\begin{proof}[Proof of Theorem~\ref{prop:Korrektheit}]
	The claim follows from combining Lemmas~\ref{lem:offspring-meets-lbs} and \ref{lem:satisfaction-upper-bounds}.
\end{proof}

\begin{proof}[Proof of Proposition~\ref{prop:coverage-search-space}]
We describe a construction of $G$ and $H$, starting from $O$. For this,
we describe, how to obtain the subgraphs $\Gsub$ and $\Hsub$
from $O$, along with the crossover point $\CP$.

To obtain the subgraphs, we first choose a subgraph $\Osub$ to
act as our $\Gsub$ from our algorithm. The subgraph $\Osub$
must be feasible, so specifically all lower bounds are satisfied.
All elements in $O$ not chosen for $\Osub$ are part of $\Hsub$.
If an edge from $\Hsub$ connects to a node in $\Osub$, we make
the node in $\Osub$ part of the crossover point.

We can always obtain the resulting structures from our algorithm.
It should be obvious to see how we could obtain $\Osub$, or in
that direction $\Gsub$, as a subgraph of a larger graph $G$. Given
the appropriate $H$, we can then choose to add exactly those nodes
to the crossover point, that we included in the crossover point in our
construction.

This construction delivers subgraphs $\Gsub$ and $\Hsub$, along
with a crossover point $\CP$. Since $\Gsub$ already is feasible,
we do not need to, but can change it further to achieve $G$, for
$\Hsub$ we might need to add elements to ensure feasibility. We
can always create a feasible $H$ to $\Hsub$, the existence of
such a graph is shown by the existence of $O$.
\end{proof}